\documentclass[runningheads]{llncs}
\usepackage{graphicx}
\usepackage{algorithm}
\usepackage{algpseudocode}
\usepackage{amsmath}
\usepackage{microtype}
\usepackage{caption}
\usepackage{subcaption}
\usepackage{booktabs} 
\usepackage{amssymb}
\DeclareMathOperator*{\argmax}{arg\,max}

\usepackage{hyperref}

\usepackage{appendix}
\begin{document}
\title{Hyperboost: Hyperparameter Optimization by Gradient Boosting surrogate models.}
\titlerunning{Hyperboost}
%
\author{Jeroen van Hoof\inst{1} \and
Joaquin Vanschoren\inst{1}
}
\authorrunning{van Hoof and Vanschoren}
%
\institute{Eindhoven University of Technology, Eindhoven, The Netherlands}
\maketitle              
\begin{abstract}
Bayesian Optimization is a popular tool for tuning algorithms in automatic machine learning (AutoML) systems. Current state-of-the-art methods leverage Random Forests or Gaussian processes to build a surrogate model that predicts algorithm performance given a certain set of hyperparameter settings. In this paper, we propose a new surrogate model based on gradient boosting, where we use quantile regression to provide optimistic estimates of the performance of an unobserved hyperparameter setting, and combine this with a distance metric between unobserved and observed hyperparameter settings to help regulate exploration. We demonstrate empirically that the new method is able to outperform some state-of-the art techniques across a reasonable sized set of classification problems.

\end{abstract}
\section{Introduction}\label{sec:intro}
Hyperparameter optimization (HPO) is a common problem in machine learning. Machine learning algorithms, from logistic regression to neural nets, depend on well-tuned hyperparameters to reach maximum effectiveness \cite{kohavi1995automatic}\cite{mantovani2016hyper}\cite{olson2017data}\cite{sanders2017informing}\cite{thornton2013auto}.

However, optimizing hyperparameters by hand requires a lot of manual programming and testing, as well as considerable experience with machine learning algorithms. By automating this process, we can assist the user in finding better solutions faster, and often find solutions that outperform manually tuned models.

Instead of resorting to standard techniques like grid search or random search, Bayesian optimization offers an alternative and often very effective technique to tackle the problem of hyperparameter optimization. It recently gained traction in HPO by obtaining new state-of-the-art results in tuning deep neural networks for image classification \cite{snoek2012practical}\cite{snoek2015scalable}, speech recognition \cite{dahl2013improving} and neural language modeling \cite{melis2017state}, and by demonstrating wide applicability to different problem
settings.

Bayesian optimization (BO) is an iterative algorithm with two key ingredients: a probabilistic surrogate model and an acquisition function to decide which point to evaluate next. In each iteration, the surrogate model is fitted to all observations of the target function made so far. Then the acquisition function, which uses the predictive distribution of the probabilistic model, determines the utility of different candidate points, trading off exploration and exploitation. Compared to evaluating the expensive target function, the acquisition function is cheap to compute and can therefore be thoroughly optimized. In the case of HPO, the target function is a machine learning model (or pipeline) we want to optimize. We will describe BO in more detail in \autoref{subsec:bo}.

Two main machine learning models used as a probabilistic surrogate model are Gaussian Processes (GPs) and Random Forests. While GPs perform better than Random Forests on low-dimensional, numerical configuration spaces \cite{eggensperger2013towards}, Random Forests natively handle high-dimensional, partly discrete and conditional configuration spaces better where standard GPs do not work well \cite{eggensperger2013towards}\cite{jenatton2017bayesian}\cite{li2016hyperband}.


Furthermore, the training time of GPs grows cubically with the number of training samples and hyperparameters, whereas the runtime of tree-based models only grow linearly with the number of samples. The time needed to select promising configurations for BO tools based on GPs therefore increases much faster as more configurations are observed.

Current state-of-the-art BO tools for HPO, such as the Sequential Model-based Algorithm Configuration (SMAC)~\cite{hutter2011sequential} library, use the Random Forest model as a surrogate model. However, in machine learning problems, it has often been observed that gradient boosting decision tree (GBDT) models provide better performance than Random Forests \cite{olson2017data}.

We propose a new surrogate model based on gradient boosting. We will show however, that it is non-trivial to obtain uncertainty measures from a gradient boosting model. We therefore propose an alternative, novel approach based on the following ingredients:
\begin{itemize}
\item An optimistic estimate of an unseen candidate configuration based on quantile regression using the gradient boosting model.
\item An estimate of uncertainty for an unseen candidate configuration caused by sparsity in the distribution of observed configurations. This estimate is calculated independently of the model.
\item An acquisition function that combines the aforementioned metrics into one utility metric to quantify promising configurations.
\end{itemize}

We will call our new algorithm Hyperboost, where the ``hyper" refers to hyperparameters and ``boost" to gradient boosting.




We will now provide an overview of the structure of the paper. We will cover related work in such as SMAC and the Tree Parzen Estimator algorithm as well as the types of uncertainty that are important for Bayesian Optimization in \autoref{sec:related-work}. We then go more in depth on the topic of Bayesian Optimization and acquisition functions and give an introduction to gradient boosting in \autoref{sec:preliminaries}. Then, in \autoref{sec:hyperboost}, we describe the challenges of adapting Gradient Boosting to Bayesian Optimization and lay out the ingredients of Hyperboost. We continue by describing our methods (\autoref{sec:methods}), experiments (\autoref{sec:experiments}) and results (\autoref{sec:results}). And finally, we end with a conclusion (\autoref{sec:conclusion}) and future work (\autoref{sec:futurework}).

\section{Related work}\label{sec:related-work}

In \autoref{subsec:surrogates} we cover the evolution and related work in surrogate models used for Bayesian optimization and explain how AutoML tools like SMAC use Random Forests to perform efficient Bayesian Optimization. In \autoref{subsec:tpe} we discuss the Tree Parzen Estimator algorithm that uses Kernel Density Estimators to model the distributions of good and bad observations. Lastly, in \autoref{subsec:uncertainties} we describe two types of uncertainties and how they relate to quantile regression (which we use to create optimistic estimates) and density (which we use to apply preference to sparse regions).

\subsection{Surrogate models}\label{subsec:surrogates}
Traditionally, Bayesian optimization employs Gaussian processes \cite{williams2006gaussian} to model the target function because of their expressiveness, smooth and well-calibrated uncertainty estimates and closed-form computability of the predictive distribution.

However, as mentioned in \autoref{sec:intro}, Gaussian Processes scale poorly in the number of samples and number of dimensions. More specifically, the computational complexity of fitting and predicting variances with GPs for $n$ data points scales as $O(n^3)$ and $O(n^2)$. Therefore, a large amount of research is done on extending GPs to efficiently handle configuration spaces with a larger number of hyperparameters \cite{wang2016bayesian}\cite{wang2017batched}\cite{gong2019quantile}\cite{gardner2017discovering}\cite{kandasamy2015high} and on adapting more scalable and flexible machine learning models such as neural networks \cite{snoek2015scalable}\cite{springenberg2016bayesian}.

Another alternative model for Bayesian optimization are Random Forests \cite{hutter2010sequential}. Where GPs perform better than Random Forests on small, numerical configuration spaces, Random Forests natively handle larger, categorical and conditional configuration spaces. The computational complexity of fitting and predicting variances for Random Forests, is only $O(n \log n)$ and $O(\log n)$, respectively.

Random Forests are used for the surrogate model by SMAC. Normally, Random Forests do not provide a confidence interval for their estimates. So these models need to be adapted to express uncertainty. One common approach in constructing the predictive distribution is to assume a Gaussian $\mathcal{N} (\hat{\mu}, \hat{\sigma}^2)$.

The first version of SMAC~\cite{hutter2011sequential} initially computed the Random Forest’s predictive mean $\hat{\mu}_{\boldsymbol{x}}$ and variance $\hat{\sigma}_{\boldsymbol{x}}^2$
as the empirical mean and variance of its individual trees’ predictions for $\boldsymbol{x}$.

However, in a later paper about algorithm runtime prediction ~\cite{hutter2014algorithm}, they revised their Random Forest model and this model is used in later versions of SMAC. The new method stores, for each leaf of the regression tree, the empirical mean and empirical variance of the training data associated with that leaf.

To avoid making deterministic predictions for leaves with few data points, we round the stored variance up to at least the constant $\sigma^2_{min}$; we set $\sigma^2_{min} = 0.01$ throughout. Since each input for a regression tree will guide you to a different leaf, we can see that for any input, each regression tree $T_b$ thus yields a predictive mean $\mu_b$ and a predictive variance $\sigma^2_b$.

To combine these estimates into a single estimate, we treat the forest as a mixture model of $B$ different models. We can then compute $\hat{\mu}$ and $\hat{\sigma}^2$ as follows:

\begin{align}
\hat{\mu} &= \frac{1}{B} \sum_{b=1}^{B}{\mu_b} \\
\hat{\sigma}^2 &= \frac{1}{B} \left(\sum_{b=1}^{B}{\sigma^2_b + \mu^2_b}\right) - \mu^2
\end{align}

Here, the mean prediction is calculated as the mean of the means and the variance estimate is calculated using the law of total variance: the mean of the variance plus the variance of the means.

\subsection{Tree-structured Parzen Estimator}\label{subsec:tpe}
A particularly interesting algorithm related to Bayesian Optimization is the Tree-structured Parzen Estimator (TPE~\cite{bergstra2011algorithms}\cite{bergstra2013making}). What makes this algorithm relevant, is that it makes use of density, which is also a key ingredient of our algorithm.

In contrast to normal Bayesian Optimization, which models the probability $p(y|\boldsymbol{x})$ of observations $y$ given the configurations $\boldsymbol{x}$, the TPE  models density functions $p(\boldsymbol{x}|y < \alpha)$ and $p(\boldsymbol{x}|y \geq \alpha)$. Given a percentile $\alpha$ (usually set to 15\%), the observations are divided in good observations and bad observations and simple $1-d$ Parzen windows are used to model the two distributions. The ratio $\frac{p(\boldsymbol{x}|y < \alpha)}{p(\boldsymbol{x}|y \geq \alpha)}$ is related to the expected improvement acquisition function and is used to propose new hyperparameter configurations. 

TPE uses a tree of Parzen estimators for conditional hyperparameters and demonstrated good performance on such structured HPO tasks \cite{bergstra2011algorithms}\cite{bergstra2013making}\cite{eggensperger2013towards}\cite{falkner2018bohb}\cite{sparks2015automating}\cite{thornton2013auto}\cite{zhang2016flash}, is conceptually simple, and parallelizes naturally \cite{loshchilov2016cma}. It is also the workhorse behind the AutoML framework Hyperoptsklearn \cite{komer2014hyperopt}.

\subsection{Types of uncertainty}\label{subsec:uncertainties}
In Bayesian modeling, there are two main types of uncertainty one can model \cite{der2009aleatory}. \textbf{Aleatoric uncertainty} captures noise inherent in the observations. This type of uncertainty arises due to hidden variables or measurement errors, and cannot be reduced by collecting more data under the same experimental conditions.  On the other hand, \textbf{epistemic uncertainty} describes the errors associated to the lack of experience of our model at certain regions of the feature space. This uncertainty can be explained away given enough data, and is often referred to as model uncertainty. 

In \cite{tagasovska2018frequentist}, the authors use quantile regression to estimate aleatoric uncertainty for deep learning. They show that their estimator based on simultaneous quantile regression shows state-of-art results in providing well-calibrated prediction intervals.

The TPE algorithm described in \autoref{subsec:tpe} models two density functions over the observations. Such a density functions is tightly related to epistemic uncertainty: if we were to model a density function $p(\boldsymbol{x})$ given the configurations $\boldsymbol{x}$, epistemic uncertainty is inversely proportional to the density of  $p(\boldsymbol{x})$, and could be reduced by collecting data in those low density regions. The ratio $\frac{p(\boldsymbol{x}|y < \alpha)}{p(\boldsymbol{x}|y \geq \alpha)}$ then makes a trade-off between ``good" regions and sparse regions.

\section{Preliminaries}\label{sec:preliminaries}

In \autoref{subsec:bo}, we will give an overview of Bayesian Optimization, and cover acquisition functions in \autoref{subsec:acqfunc}. Lastly, in \autoref{subsec:lightgbm} we describe gradient boosting in general and introduce the gradient boosting framework that we will use for our experiments.

\subsection{Bayesian optimization}\label{subsec:bo}
Bayesian optimization is a powerful tool for optimizing objective
functions which are very costly or slow to evaluate. In particular, we consider problems where the maximum is sought for an expensive function $f : \mathcal{X} \rightarrow \mathbb{R}$,
$$\boldsymbol{x}_{\texttt{OPT}} = \argmax_{\boldsymbol{x} \in \mathcal{X}}{f(\boldsymbol{x})}$$ within a domain $\mathcal{X} \subset \mathbb{R}^d$ which is a bounding box.

Hyperparameter optimization for machine learning models
is of particular relevance as the computational costs for
evaluating model variations is high and hyperparameter gradients are typically not available.

Bayesian optimization techniques can be effective in practice even if the underlying function being optimized is stochastic, non-convex, or noncontinuous.

Sequential model-based optimization (SMBO) is a formalism of Bayesian optimization. The SMBO process is shown in \autoref{alg:smbo}. In this optimization process, a probabilistic regression model $M$ is initialized using a small set of samples from the domain $\mathcal{X}$. Following this initialization phase, new locations within the domain are sequentially selected by optimizing (in practice we do this with, for example, random sampling or local search) an \emph{acquisition function} $S$ which uses the current model as a cheap \emph{surrogate model} for the expensive objective function $f$. 

\begin{algorithm}[tb]
   \caption{Sequential Model-Based Optimization}
   \label{alg:smbo}
\begin{algorithmic}
   \State {\bfseries Input:} objective $f$, domain $\mathcal{X}$, acq. func. $S$, model $M$
   \State $\mathcal{D} \leftarrow$ \textsc{InitSamples}($f, \mathcal{X}$)
   \For{$i \leftarrow |\mathcal{D}|$ to $T$}
    \State $p(y|\boldsymbol{x}, \mathcal{D}) \leftarrow$ \textsc{FitModel}($M, D$)
    \State $\boldsymbol{x}_i \leftarrow \argmax_{\boldsymbol{x} \in \mathcal{X}}{S(\boldsymbol{x}, p(y | \boldsymbol{x}, D))}$
    \State $y_i \leftarrow f(\boldsymbol{x}_i)$
    \State $\mathcal{D} \leftarrow \mathcal{D} \cup (\boldsymbol{x}_i, y_i)$
   \EndFor
\end{algorithmic}
\end{algorithm}

Each suggested function evaluation produces an observed result $y_i = f(x_i)$; note that the observation may be random either because $f$ is random or because the observation process is  subject to noise. This result is appended to the historical set $\mathcal{D} = \{(x_1, y_1), \dots, (x_i, y_i)\}$, which is used to update the surrogate for generating the next suggestion. In practice, there is often a quota on the total time or resources available, which imposes a limit $T$ on the total number of function evaluations. 

\subsection{Acquisition functions}\label{subsec:acqfunc}
The acquisition function determines where to sample next by providing the expected utility of evaluating a sample $\boldsymbol{x}$. Although many acquisition functions exist, the expected improvement (EI)~\cite{jones1998efficient}:
\begin{equation}
    \mathbb{E}[\mathbb{I}(\boldsymbol{x}))] = \mathbb{E}[\max(y - f_{max}, 0)]
\end{equation}
is common choice since it can be computed in closed form if the model prediction
$y$ at configuration $\lambda$ follows a normal distribution:
\begin{equation}
    \mathbb{E}[\mathbb{I}(\boldsymbol{x})] = (\mu(\boldsymbol{x}) - f_{max}) \Phi \frac{\mu(\boldsymbol{x}) - f_{max}}{\sigma} + \sigma \phi  \frac{\mu(\boldsymbol{x}) - f_{max}}{\sigma}
\end{equation}

where $\phi(\cdot)$ and $\Phi(\cdot)$ are the standard normal density and standard normal distribution function, and $f_{max}$ is the best observed value so far.

Other examples of acquisition functions are Probability of improvement~\cite{lizotte2008practical}, Entropy search~\cite{hennig2012entropy} and Upper confidence bound~\cite{srinivas2009gaussian}.

\subsection{Gradient boosting}\label{subsec:lightgbm}
Gradient boosting is a machine learning technique that produces a prediction model in the form of an ensemble of weak prediction models, typically decision trees. It builds the model in a stage-wise fashion, and it generalizes them by allowing optimization of an arbitrary differentiable loss function.

Starting with a weak base model, models are trained iteratively, where each new model is trained on the error residuals of the previous model. In other words, each new model adds to the prediction of the previous model to produce a strong overall prediction. However, since the ensemble of models is built in a stage-wise fashion, it is not possible to build the weak learners in parallel, and as a result, early gradient boosting algorithms are not parallelized.

Later algorithms, such as XGBoost~\cite{chen2015xgboost}, do however make gradient boosting parallel, not by creating multiple decision trees in parallel, but by using parallelization within a single tree. These algorithms collect statistics for each column in parallel, which leads to a parallel algorithm for split finding. Further optimizations for XGBoost deal with sparsity and cache-awareness.

LightGBM~\cite{ke2017lightgbm} introduced Exclusive Feature Bundling (EFB), which bundles features together that are (almost) exclusive, i.e., they rarely take nonzero values simultaneously (e.g. one-hot-features). EFB merges many sparse features into much fewer features, which leads to previously isolated features to be bundled together. This can increase spatial locality (i.e. the use of data elements within relatively close storage locations) and improve cache hit rate significantly. The result is a significant speed boost over other gradient boosting algorithms, such as XGBoost. We also found that for our cases, at the time of writing, LightGBM was slightly faster than CatBoost~\cite{dorogush2018catboost}. We therefore decided on using the LightGBM implementation of gradient boosting.

\section{Hyperboost}\label{sec:hyperboost}

In this section, we will describe the main ingredients of Hyperboost. Firstly, in \autoref{subsec:limitations} we discuss the challenges and limitations of adapting Gradient Boosting to Bayesian Optimization.

In \autoref{subsec:qr} we will describe how and why we create optimistic estimates for unseen candidate configurations based on quantile regression using the gradient boosting model. Then, in \autoref{subsec:dtnn} we show how we create an estimate of uncertainty for an unseen candidate configuration caused by sparsity in the distribution of observed configurations. This estimate is based on the distance to the nearest observed configuration.

Finally, in \autoref{subsec:acqfunc2} we define the acquisition function that combines the aforementioned metric into one utility metric to quantify promising configurations.

\subsection{Limitations of Gradient Boosting}\label{subsec:limitations}

Whereas we can calculate the predictive uncertainty for Random Forests by using the empirical variance of the individual trees' predictions or the empirical variance of the training data associated with the leafs, this will not work well for gradient boosting techniques. 

In the gradient boosting decision tree (GBDT) algorithm, each new tree helps to correct errors made by the previously trained trees. This leads to the earlier trees contributing more towards the final estimate, while the later trees are correcting the earlier trees' estimates. And thus, the individual trees are not independent from one another.

Instead, we can use quantile regression to calculate a prediction interval and hence quantify some degree of uncertainty. One attempt at using quantile regression via Gradient Boosting within Bayesian optimization comes from the Scikit-optimize package \cite{scikitoptimize}. Here, they use three separate gradient boosting models: one model to estimate the median, and two models to estimate the 16th and 84th percentile, which are, in a normal distribution, roughly equivalent to a score that is 1 s.d. below and 1 s.d. above the mean. Taking the average of these two values, gives us an estimation of the standard deviation.

However, we found that this approach led to bad performance: the uncertainty estimate is only based on aleatoric uncertainty and there is no incentive to explore sparse regions. Whereas in Gaussian Processes, uncertainty around observations is decreased. For Random Forests also holds that, because of randomness in the construction of the base learners such as bagging, we have a broader distribution of predictions from the base learners for sparse regions.

\subsection{Optimistic estimates using quantile regression}\label{subsec:qr}
Quantile regression is an alternative to ordinary least squares regression. Whereas the sum of squared errors is minimized in ordinary least squares regression, the median regression estimator minimized the sum of absolute errors. The remaining conditional quantile functions are estimated by minimizing an asymmetrically weighted sum of absolute errors:

\begin{equation}
    S_1^* = \sum{i=1}^{n}{w_i(p) \| y_i - \gamma_i(p) \|}
\end{equation}

where $n$ is the sample size, $y_i$ is the response variable, $\gamma_i(p)$ is the population quantile and the weights are defined as follows:

\begin{equation}
    w_i(p) = \begin{cases}
      $p$, & \text{if}\ y_i > \gamma_i(p) \\
      $1 - p$, & \text{if} y_i \leq \gamma_i(p)
    \end{cases}
\end{equation}


Using GBDT models, it is not possible to predict more than one value per model (because we are optimizing each model for one particular loss function). So normally, two models are required to calculate the prediction interval and one to calculate the median function. In our case however, we are more interested in the upper prediction boundary. 

From this boundary we can construct an optimistic guess as to how good a certain area in the hyperparameter space will be. If our guess is wrong, then our optimistic guess will quickly decrease and we will be compelled to move to different areas in the hyperparameter space. On the other hand, if our guess is correct, we can exploit this region to find more good results.

Using these optimistic guesses alone to determine which configuration to investigate next, however, results in over-exploitation. The reason for this, is that there is no incentive to take distance from configurations that were already tried out, especially if these configurations showed good results. We therefore need to factor in this distance to help regulate our exploration.

\subsection{Distance to nearest neighbor}\label{subsec:dtnn}


We first define a vector representation $v(\boldsymbol{x}) \in \{(0,1)^{(0)}, (0,1)^{(1)}, \dots (0,1)^{(n)}\}$ for a configuration $\boldsymbol{x}$ with $n$ hyperparameters. Here, for numerical hyperparameters, 0 represents the lowest value in the parameter space, while 1 represents the highest value. The actual hyperparameter values do not have to be in this range, but rather this representation maps to hyperparameters' actual values. Categorical hyperparameters are one-hot-encoded and concatenated with the vector.

Next, we let $d(\boldsymbol{x}, \boldsymbol{y})$ be the Manhattan distance between vectors $v(\boldsymbol{x})$ and $v(\boldsymbol{y})$. Here we choose for the Manhattan distance because we want to measure the similarity of two configurations per hyperparameter individually. During training time, we construct a KD-tree over the vector representations of all observed configurations $\{v(x) | x \in \mathcal{X}\}$. Then, during prediction, for an unseen configuration $\boldsymbol{u}$ we query the tree and calculate the Manhattan distance $d(\boldsymbol{u}, \boldsymbol{u}')$ to its nearest neighbor $u'$.

We then normalize $d(\boldsymbol{u}, \boldsymbol{u}')$ by a constant $m$, which is defined as the the maximum Manhattan distance (the number of numerical hyperparameters plus the number of categories per categorical hyperparameter), to get a relative distance $0 \leq \delta_u \leq 1$.

\begin{equation}\label{eq:delta}
    \delta_{\boldsymbol{u}} = d(\boldsymbol{u}, \boldsymbol{u}') / m
\end{equation}


\subsection{Acquisition function}\label{subsec:acqfunc2}
We then construct an acquisition function that combines (1) quantile regression to provide optimistic estimates and (2) distance-to-nearest-neighbor to help regulate our exploration:

\begin{equation}\label{eq:acq}
    Acq(\boldsymbol{x}) = \hat{q}_{\boldsymbol{x}} + s * \delta_x
\end{equation}

Where:
\begin{itemize}
    \item $\hat{q}_{\boldsymbol{x}}$ is the quantile estimate for the performance of configuration $\boldsymbol{x}$, produced by the gradient boosting regression algorithm. In our case, we use the 0.90 quantile.
    \item $s$ is a scalar that determines the importance of distance. We set $s$ equal to the standard deviation of the observed configurations so far.
    \item $\delta_{\boldsymbol{x}}$ is defined in \autoref{eq:delta}.
\end{itemize}

\section{Methods}\label{sec:methods}
We modify SMAC to make use of our methods. In that way, we start with a complete HPO tool and only need to change some components. And since SMAC is also used by AutoML tools such as AutoSklearn, it would be easier to modify such tools to make use of our proposed methods. We make the following changes to SMAC:
\begin{itemize}
    \item Instead of SMAC's Random Forest, we use LightGBM's Gradient Boosting algorithm to estimate the 90th quantile.
    \item As an addition, we construct a kD-tree to determine the Manhattan distance from a candidate configuration to the nearest observed configuration.
    \item We replace the Expected Improvement acquisition function with \autoref{eq:acq}, that combines the estimate from quantile regression with the sparsity estimate resulting from the Manhattan distance into one utility metric.
\end{itemize}

By constructing and comparing learning curves of the LightGBM model on a meta-dataset of randomly selected configurations executed over a small set of datasets, we found that a LightGBM regression model with a maximum number of leaves of 8 per tree produced the best anytime prediction results. One other advantage of having a small number of leaves per tree, is that these models are relatively fast to construct. We use 100 gradient boosting iterations, whereas SMAC only builds 10 individual trees. The runtime for building the LightGBM model and constructing the KD-tree is very similar to the runtime required to construct SMAC's Random Forest.

\section{Experiments}\label{sec:experiments}

In this section we will explain briefly how we setup our experiments. Firstly, we will give an overview of the datasets in \autoref{subsec:datasets}. Then we describe the models and their configuration spaces in \autoref{subsec:config-spaces}. After that, in \autoref{subsec:data-splits} we show what part of the data is used for optimizing the hyperparameters, and what part is used for validating the final result. We will also explain how this data is handled internally by the AutoML algorithm to try out different configurations. Finally, we describe how we configured SMAC and Hyperboost in \autoref{subsec:configuration}.

\subsection{Datasets}\label{subsec:datasets}
We used a subset of 31 datasets from the OpenML CC18\footnote{\url{https://www.openml.org/s/99}} benchmarking suite. To reduce computation time, we dropped a few very large datasets, with the addition of some randomly selected smaller datasets. The full list of tasks used in these experiments can be found in \autoref{tab:task-overview-lightgbm}.

\subsection{Configuration spaces}\label{subsec:config-spaces}
We conduct our experiments using three different models with each their own configuration space. We will use the following models:
\begin{itemize}
\item A Random Forest model from LightGBM's framework.
\item A Support Vector Machine from the Scikit-Learn package, which is in turn based on Liblinear~\cite{fan2008liblinear}.
\item A Decision Tree model from the Scikit-Learn package.
\end{itemize}

The configuration spaces of these models can be seen in \autoref{tab:random-forest}, \autoref{tab:decision-tree} and \autoref{tab:svm} respectively. 

\begin{table}[]
\centering
\begin{tabular}{llll}
\hline
\textbf{Parameter}  & \textbf{Type} & \textbf{Range}   & \textbf{Default} \\ \hline
colsample\_bytree   & float         & {[}0.20, 0.80{]} & 0.70             \\
subsample           & float         & {[}0.20, 0.80{]} & 0.66             \\
num\_leaves         & integer       & {[}4, 64{]}      & 32               \\
min\_child\_samples & integer       & {[}1, 100{]}     & 20               \\
max\_depth          & integer       & {[}4, 12{]}      & 12               \\ \hline\\
\end{tabular}
\caption{Parameter space of Random Forest.\\}
\label{tab:random-forest}

\begin{tabular}{llll}
\hline
\textbf{Parameter}  & \textbf{Type} & \textbf{Range}          & \textbf{Default} \\ \hline
criterion           & cat.   & gini, entropy & gini           \\
max\_depth          & integer       & {[}1, 20{]}             & 20               \\
min\_samples\_split & integer       & {[}2, 20{]}             & 2                \\
min\_samples\_leaf  & integer       & {[}1, 20{]}             & 1                \\ \hline\\
\end{tabular}
\caption{Parameter space of Decision Tree.\\}
\label{tab:decision-tree}

\begin{tabular}{llll}
\hline
\textbf{Parameter} & \textbf{Type} & \textbf{Range}       & \textbf{Default} \\ \hline
tol                & log. float    & {[}1e-5, 1e-1{]}     & 1e-4             \\
C                  & log. float    & {[}0.03125, 32768{]} & 1.0              \\
\hline\\
\end{tabular}
\caption{Parameter space of SVM.}
\label{tab:svm}
\end{table}

\subsection{Data splits}\label{subsec:data-splits}
\begin{figure}
    \centering
    \includegraphics[width=0.4\linewidth]{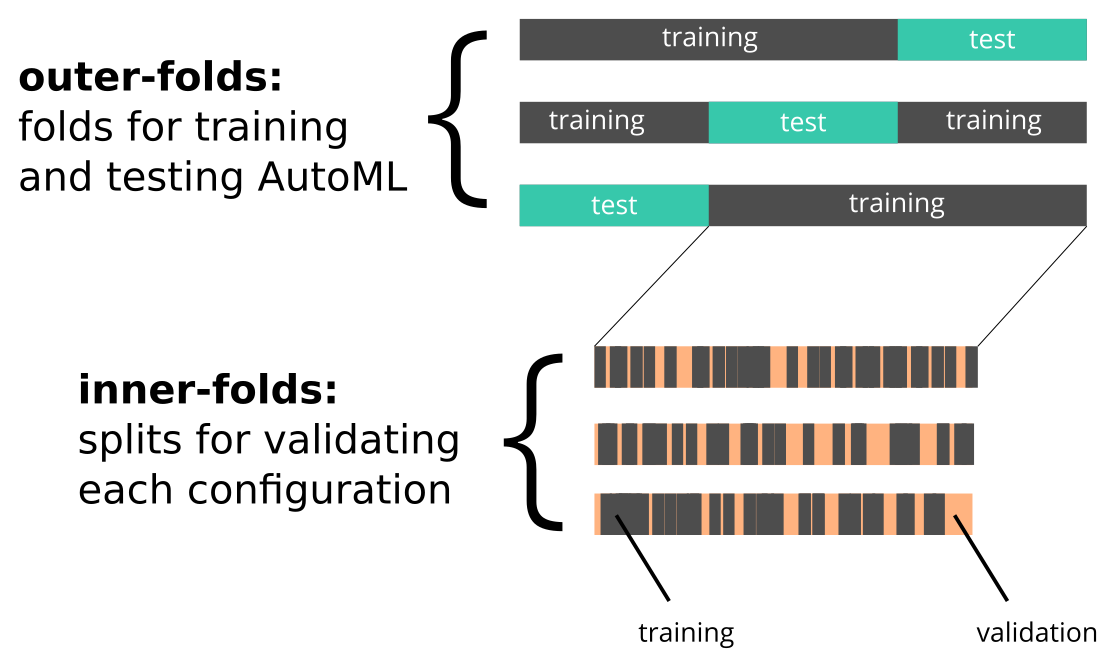}
    \caption{An overview of the data splits used for training and testing AutoML algorithms.}
    \label{fig:folds}
\end{figure}

For every target model, we repeat our experiments 3 times per dataset. For each AutoML algorithm, we furthermore execute the optimization loop once for each outer-fold as shown in \autoref{fig:folds}. Each outer-fold consists of 2/3th training data to run the AutoML algorithm over and 1/3th test data to validate the resulting model after optimization. This resulting model is first retrained on the training data and its accuracy is then validated on the test data. We repeat the process of retraining and validating 3 times for models that are build non-deterministically, such as Random Forests, and take the average accuracy of each run to get a better estimate of the performance.

When optimizing the model, we use use three inner-folds to try each configuration, which is also shown in \autoref{fig:folds}. More specifically, we shuffle the data 3 times such that we have 3 random permutations of the data and then split each permutation into 90\% training data and 10\% test data. We keep these splits the same for every configuration we try.

\subsection{Configuration}\label{subsec:configuration}
Both SMAC and Hyperboost are configured to handle non-deterministic target algorithms. This makes the SMAC framework decide on how many reruns of a configuration are necessary to determine the new best configuration. We set the maximum number of reruns per configuration to 5.

\section{Results}\label{sec:results}

In this section we will compare the performance of SMAC and Hyperboost against a baseline in \autoref{subsec:performance}. Then in \autoref{subsec:surrogates-overhead} we measure and compare the overhead times required for training and querying the surrogate models (SMAC's Random Forest and our LightGBM instance). And lastly, in \autoref{subsec:kdtree-overhead} we show the times required to construct and query the kD-Tree, which we use to calculate our distance metric. We note that the kD-Tree suffers from the curse of dimensionality and suggest one way to solve this.

\subsection{Performance}\label{subsec:performance}
\begin{figure}
  \begin{subfigure}{\linewidth}
  \includegraphics[width=0.5\linewidth]{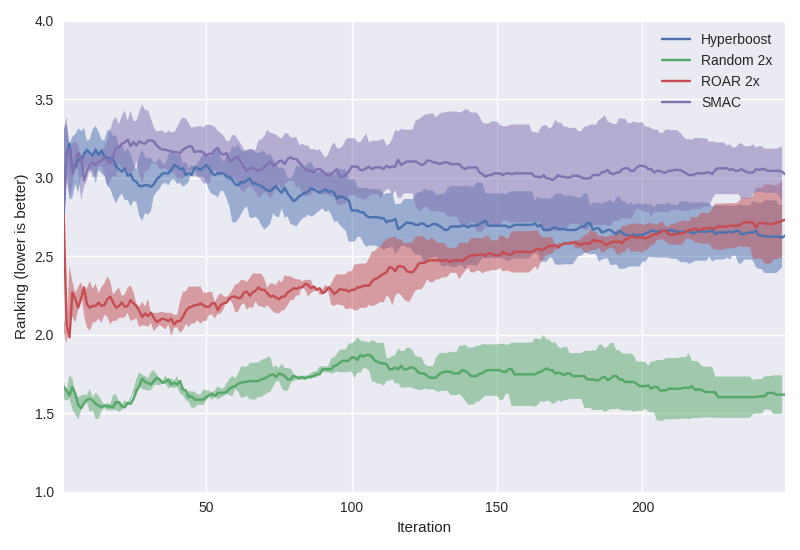}\hfill
  \includegraphics[width=0.5\linewidth]{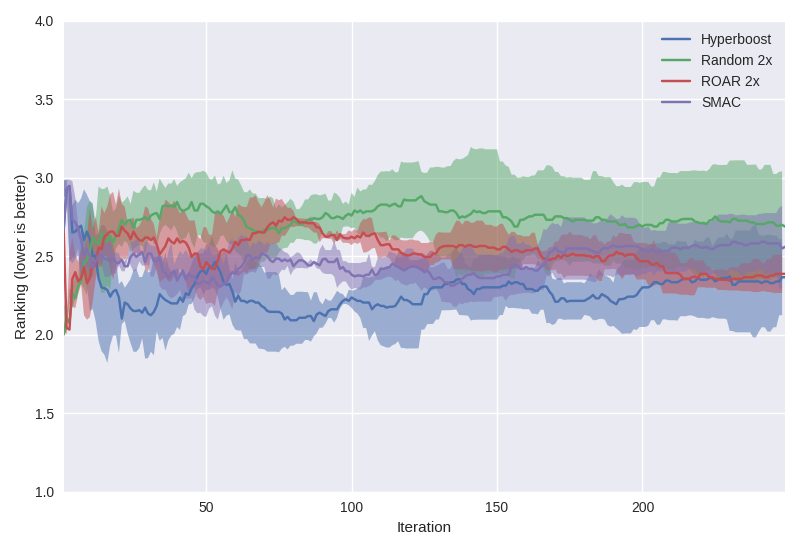}\hfill
  \caption{Ranking on training set (left) and test set (right) for optimizing the Decision Tree target model}
  \label{fig:dt}
  \end{subfigure}\par\medskip
  \begin{subfigure}{\linewidth}
  \includegraphics[width=0.5\linewidth]{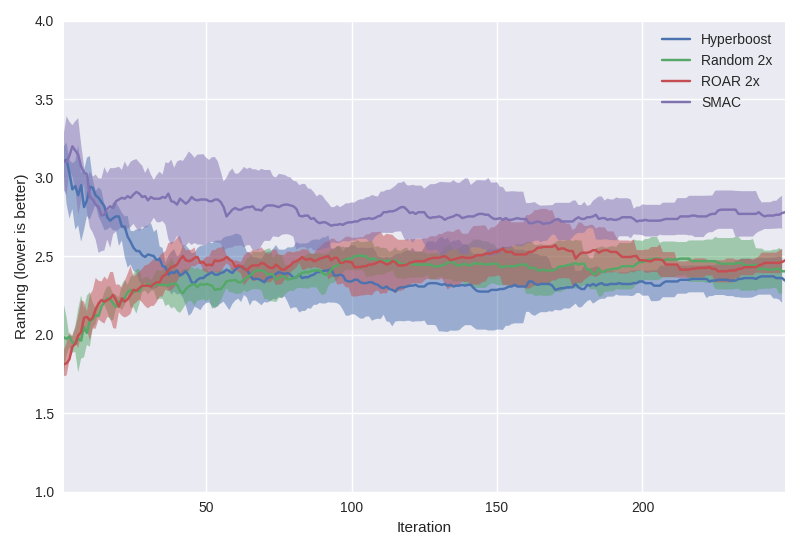}\hfill
  \includegraphics[width=0.5\linewidth]{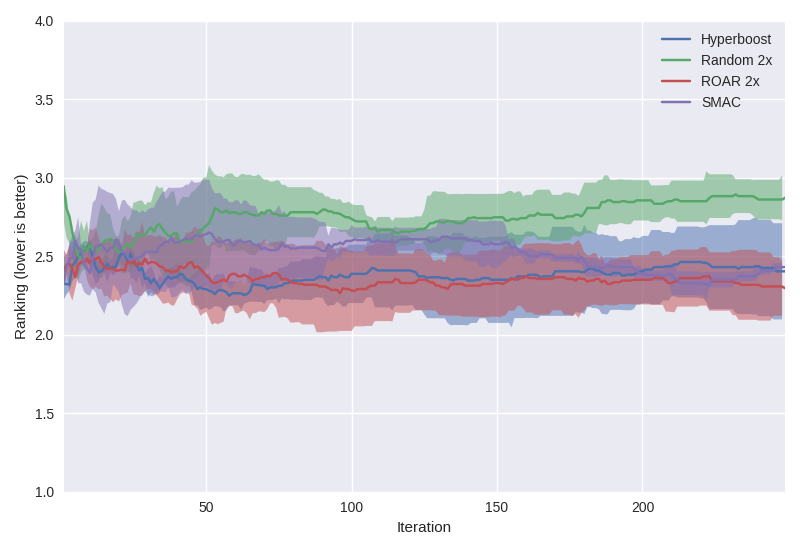}\hfill
  \caption{Ranking on training set (left) and test set (right) for optimizing the SVM target model}
  \label{fig:svm}
  \end{subfigure}\par\medskip
  \begin{subfigure}{\linewidth}
  \includegraphics[width=0.5\linewidth]{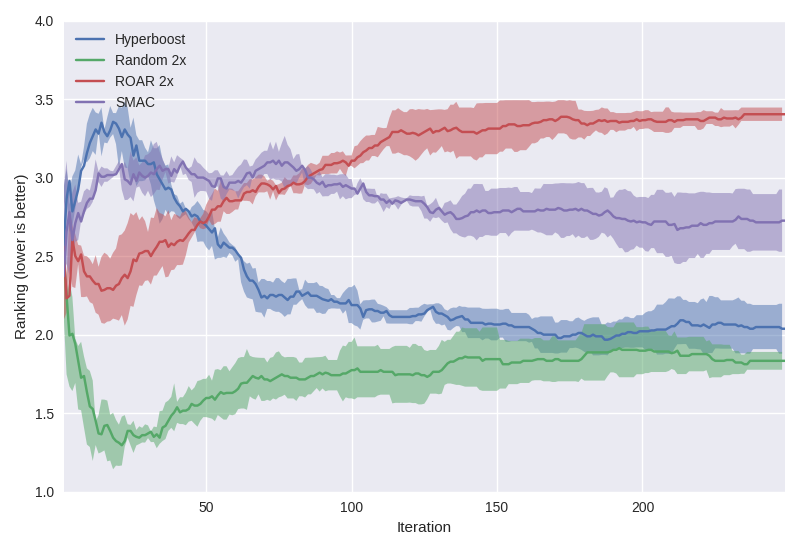}\hfill
  \includegraphics[width=0.5\linewidth]{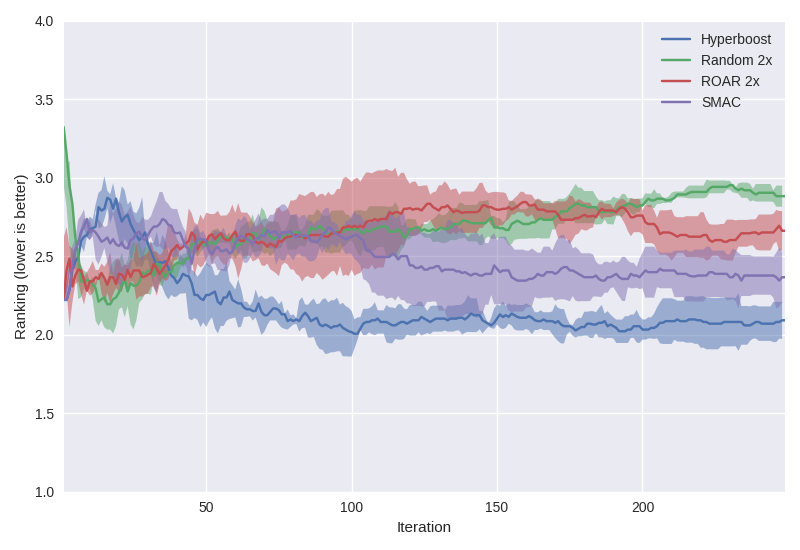}\hfill
  \caption{Ranking on training set (left) and test set (right) for optimizing the Random Forest target model}
  \label{fig:rf}
  \end{subfigure}
\end{figure}

We show the average ranking of each algorithm in \autoref{fig:dt}, \autoref{fig:svm} and \autoref{fig:rf}. This average ranking is determined as follows:
\begin{itemize}
    \item For each dataset, we calculate the ranking between algorithms on that dataset, for each optimization step.
    \item We then take the average of these rankings over all datasets.
    \item We repeat the previous steps 3 times, as we have repeated our experiments 3 times per dataset. So we now have 3 rankings per algorithm per optimization step.
    \item Finally, we take the average over the 3 rankings, such that we have one ranking per algorithm per optimization step: these are the lines in the figures. We furthermore take the standard deviation over the 3 rankings, which is shown in the shaded area of the figures.
\end{itemize}

Next to the average ranking, we show the average loss of each algorithm in \autoref{fig:dt-mean}, \autoref{fig:svm-mean} and \autoref{fig:rf-mean}. The average loss is calculated by taking the average over all datasets and repetitions.

The figures show that Hyperboost performs better than SMAC in optimizing the target algorithm on the training data in every case. Of course, we hope that by optimizing the target algorithm on the training set, the resulting configuration will also work well on the test set. The figures \autoref{fig:dt-mean}, \autoref{fig:svm-mean} and however clearly show that overfitting finds place, while for a larger hyperspace, \autoref{fig:rf-mean} we can still see some improvement over a longer period of optimization.

We can furthermore see that randomized search with double the amount of resources (Random 2x) outperforms Hyperboost and SMAC on the training dataset, at least in the beginning. We can also observe that over time, the advantage of random search becomes weaker. It is furthermore important to note that randomized search runs each sampled configuration only once, whereas SMAC and Hyperboost run each configuration up to five times to get a more accurate estimation of its performance. This makes it easier for randomized search to encounter high-performing runs during training time, but that does not guarantee that this configuration shows best performance on average. That is also the reason why randomized search shows the worst performance on the test set.

It is therefore more interesting to compare against ROAR with double the amount of resources (ROAR x2). ROAR stands for Random Online Aggressive Racing and is the basis of the SMAC framework. The only difference with SMAC is that configurations are chosen at random rather than by optimizing an acquisition function. Like SMAC and Hyperboost, we have configured ROAR to run each configuration up to five times.

Looking at the training results, ROAR x2 shows initally better performance and is eventually surpassed by Hyperboost. It is likely that a higher number of hyperparameters gives Hyperboost and SMAC the advantage as these can limit their resources to areas it perceives to be interesting, while ROAR uniformly distributes it resources and hence distributes more resources over hyperparameters that are less important for the performance of the machine learning algorithm.

\subsection{Overhead times of surrogate models}\label{subsec:surrogates-overhead}
We are also interested in the time required to retrain the surrogate model and selecting the  most promising configuration. However, quantifying the overhead differences between SMAC and Hyperboost comes with a set of challenges:

\begin{itemize}
    \item Firstly, SMAC's surrogate model is implemented as a single-threaded Random Forest, whereas LightGBM can create the nodes within its trees in parallel and GPU acceleration is supported as well.
    \item Secondly, both surrogate models scale differently with the number of samples as well as the number of features.
    \item Thirdly, the overhead is divided over multiple iterations due to SMAC's racing mechanism that is used in both SMAC and Hyperboost. This racing mechanism determines how many times a configuration should be tested: each repeated run counts as an additional iteration, but does not require the surrogate model to be retrained and queried.
    \item Fourthly, SMAC (and thus Hyperboost) use the surrogate model to predict over a large batch of randomly sampled configurations and additionally uses Local Search starting from a small set of samples. The batch size and the number of Local Search steps performed are factors that also play a role in the overhead times.
\end{itemize}
 
To get a general idea of the overhead differences between SMAC and Hyperboost, we remove the racing mechanism (i.e. we evaluate each selected configuration only once) and disable Local Search. In each iteration, we simply train the model over the observed configurations and predict over a batch of 10.000 randomly sampled configurations. At the end of the iteration, the most promising configuration is evaluated and added to the list of observed configurations.

\begin{figure}
  \begin{subfigure}{\linewidth}
  \includegraphics[width=0.5\linewidth]{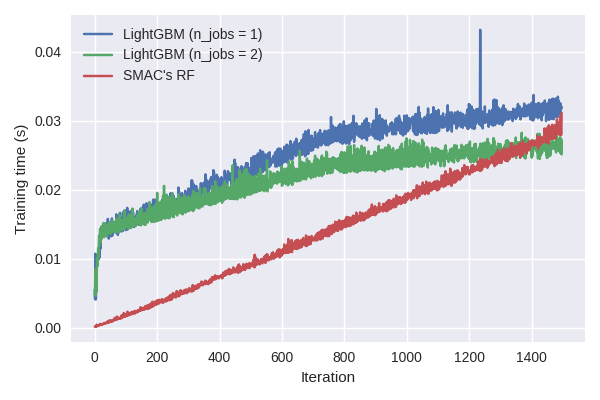}\hfill
  \includegraphics[width=0.5\linewidth]{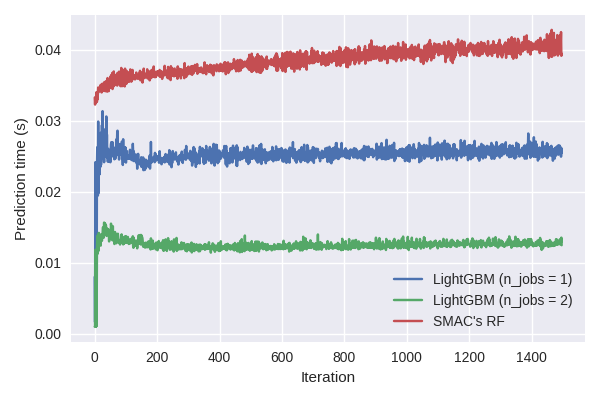}\hfill
  \caption{Training times (left) and prediction times (right) for optimizing 2 hyperparameters.}
  \label{fig:overhead1}
  \end{subfigure}\par\medskip
  \begin{subfigure}{\linewidth}
  \includegraphics[width=0.5\linewidth]{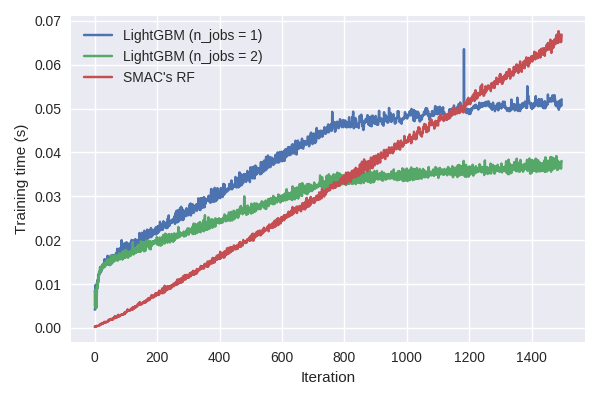}\hfill
  \includegraphics[width=0.5\linewidth]{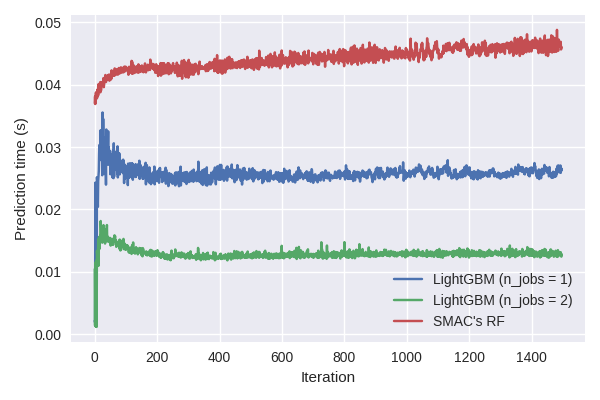}\hfill
  \caption{Training times (left) and prediction times (right) for optimizing 4 hyperparameters.}
  \label{fig:overhead2}
  \end{subfigure}
\end{figure}

\autoref{fig:overhead1} and \autoref{fig:overhead2} show the average (the average over three repetitions) training and prediction times for SMAC's Random Forest model and our LightGBM instance. We show the results for optimizing 2 and 4 hyperparameters respectively. For LightGBM, we additionally show the resulting times when using 1 and 2 threads.

Looking at the training times, we can see that LightGBM scales much better for a larger amount of samples. There is also a visible kink in the graph at the 800th iteration. This might have to do something with the fact that we are building 100 trees with a maximum of 8 leaves per tree. Also note that SMAC's Random Forest builds only 10 trees (with no maximum depth).

For the prediction times, we can also see that our LightGBM instance has much faster prediction times. LightGBM efficiently handles a large batch of samples in parallel and due to the set maximum number of leaves, the lookup times per tree are kept short. We can see however, that the prediction time for SMAC's Random Forest are growing because the trained model is still allowed to grow deeper.


\subsection{Overhead times of constructing and querying a kD-Tree}\label{subsec:kdtree-overhead}
Additionally to the overhead times of the surrogate models, we need to count the runtime required for constructing and querying a kD-Tree. The variable $k$ stands for the number of dimensions, which is equal to the number of hyperparameters we are optimizing. 

\begin{figure}
    \begin{subfigure}{\linewidth}
  \includegraphics[width=0.5\linewidth]{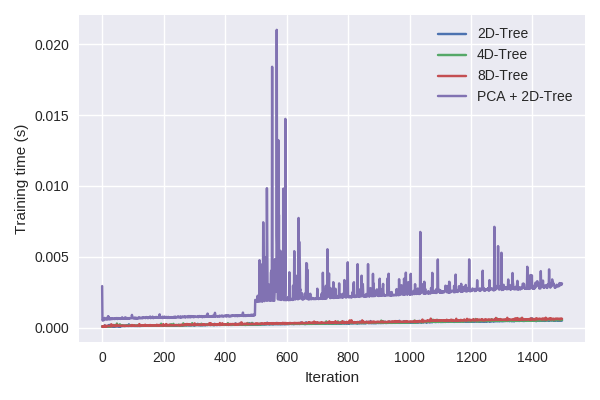}\hfill
  \includegraphics[width=0.5\linewidth]{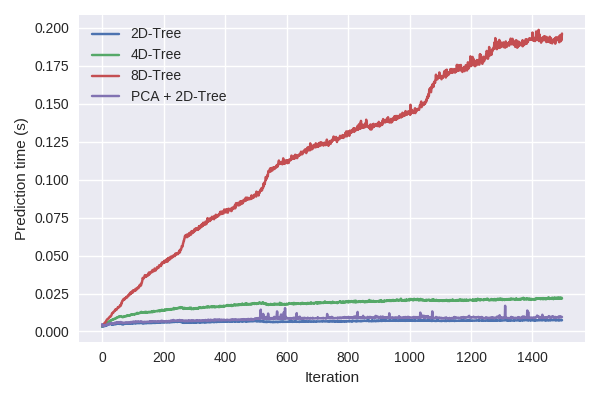}\hfill
  \caption{Construction times (left) and query times (right) for constructing and querying a $k$D-Tree. The graphs include a run where we use PCA to reduce the number of dimensions from 8 to 2 first and then construct a 2D-Tree.}
  \label{fig:overhead4}
  \end{subfigure}\par\medskip
\end{figure}

\autoref{fig:overhead4} shows these overhead times for a 2D-Tree, 4D-Tree and 8D-Tree. As it turns out, the training times are negligible in comparison to the overhead times of the surrogate model, however, the prediction time on a large batch of 10.000 samples rises very quickly for higher dimensions.

This clearly shows the curse of dimensionality which leads to most searches in high dimensional spaces to end up being needlessly complicated brute searches. As a general rule, if the dimensionality is $k$, the number of points in the data $N$, should be $N \gg 2^k$. Otherwise, when kD-Trees are used with high-dimensional data, most of the points in the tree will be evaluated and the efficiency is no better than exhaustive search~\cite{toth2017handbook}.

In the case of high-dimensional data (i.e. when we optimize over many hyperparameters), we can use approximate nearest-neighbour methods to decrease the overhead times for finding close-by points of possible candidate configurations. The kD-Tree natively handles this by setting a value $\epsilon$. The the k-th returned value is guaranteed to be no further than $1 + \epsilon$ times the distance to the real k-th nearest neighbor.

\begin{figure}
    \begin{subfigure}{\linewidth}
        \includegraphics[width=0.33\linewidth]{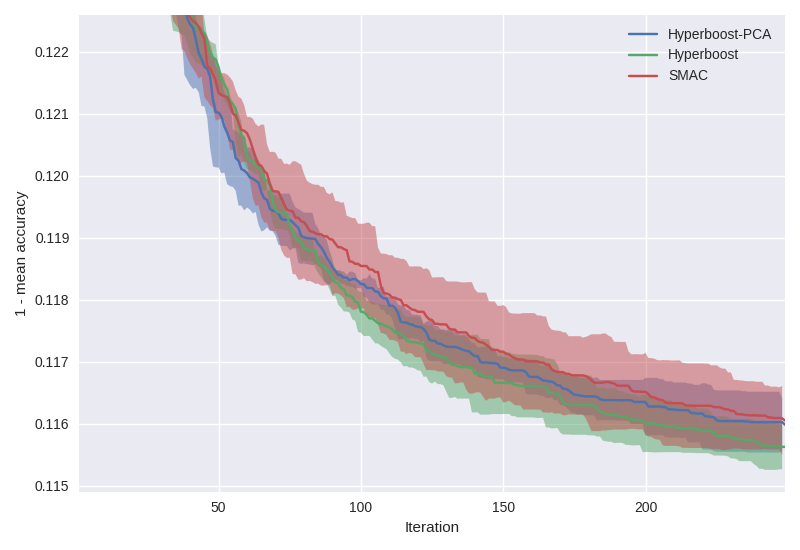}\hfill
        \includegraphics[width=0.33\linewidth]{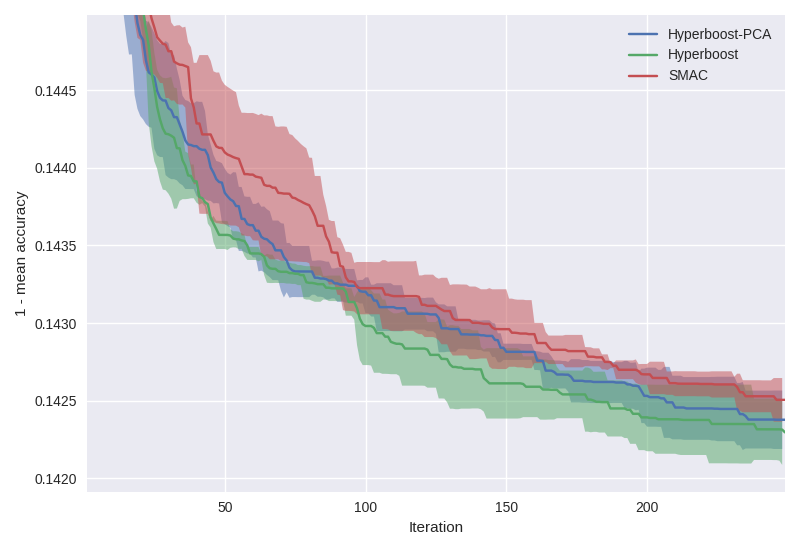}\hfill
        \includegraphics[width=0.33\linewidth]{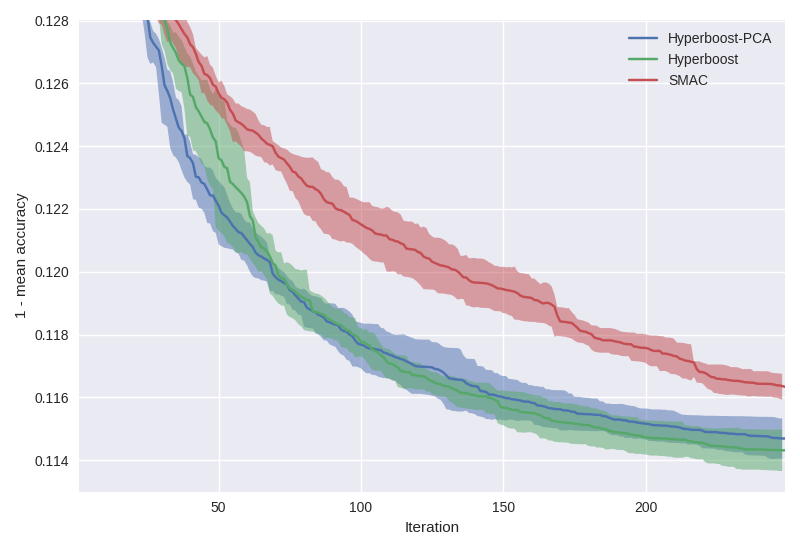}\hfill
        \caption{Effect of using PCA in combination with the KD-Tree on the Decision Tree (left), SVM (middle) and Random Forest (right) target algorithms.}
        \label{fig:pca-results}
    \end{subfigure}\par\medskip
\end{figure}

Another possibility is to map high dimensional space to a 2-dimensional space using random projection or Principle Component Analysis (PCA). These methods preserve the relative distance well. \autoref{fig:overhead4} shows the overhead of fitting and mappisng 8-dimensional data to 2-dimensional data and then constructing a 2D-tree and the overhead of mapping and querying a large batch of 10.000 samples. \autoref{fig:pca-results} shows the effect on performance of using PCA. At the cost of a small penalty in performance, we gain a large difference in overhead speed.

\section{Conclusion}\label{sec:conclusion}

In this paper, we introduced Hyperboost, an AutoML tool based on SMAC. We described the challenges of adapting the Gradient Boosting model to Bayesian Optimization, and we proceeded to adapt this model to Bayesian Optimization by making use of quantile regression. We discussed however, that using quantile regression alone is not good enough, as it does not take the epistemic uncertainty into account. By estimating the 90th quantile, we provided optimistic estimates for candidate configurations, and we combined these estimates with a bonus for the Manhattan distance to the nearest (i.e. most similar) observed configuration. The latter gives incentive to take distance from previously observed configurations. 

We demonstrated that Hyperboost was able to outperform SMAC on a reasonable set of classification problems in combination with a small set of configuration spaces. With this work, we hope to lead the way to a better alternative to SMAC with improved performance and to spark interest for gradient boosting based surrogates. However, more investigation and experimenting is required to explore the possibilities and the strengths (and weaknesses) of this combination of new ingredients for Bayesian Optimization.



\section{Future work}\label{sec:futurework}
We would like to expand our empirical tests to cover other and larger configuration spaces as well, and to increase the number of iterations and the number of datasets we experiment with. For configuration spaces, handling the CASH problem may be particularly interesting, since that enables us to test performance against tools such as (vanilla) AutoSklearn.

We would also like to include more methods to test our performance against, for example Hyperband~\cite{li2016hyperband}. 
We have currently left Hyperband out of scope because of the following reasons:
\begin{itemize}
    \item Hyperband is difficult to compare with other methods in terms of number of iterations. The iterations in Hyperband have a different budget allocated per bracket.
    \item The type of budget and the initial budget allocation for Hyperband must be carefully selected in such a way that we can speak of a fair comparison between BO-based methods and Hyperband.
    \item Where SMAC and Hyperboost update their incumbent regularly, Hyperband only produces its results in the last bracket, making it impossible to draw a curve.
    \item A comparison based on time requires a stable environment and the infinite horizon version of Hyperband.
\end{itemize}

It is desired however to include such a comparison with Hyperband, since it has interesting properties such as very good performance in the beginning~\cite{falkner2018bohb}.


We can furthermore reduce the problem of overfiting in AutoML by employing a different shuffling of the train and validation split for each function evaluation; this was shown to improve generalization performance for SVM tuning, both with a holdout and a cross-validation strategy \cite{levesque2018bayesian}.

One point of interest we have not covered, are multi-fidelity techniques. Multi-fidelity techniques can be combined with Bayesian Optimization methods, for example by using predictive termination~\cite{domhan2015speeding}, where a learning curve model is used to extrapolate a partially observed learning curve for a configuration, and the training process is stopped if the configuration is predicted to not reach the performance of the best model trained so far in the optimization process.

We also think that our acquisition function could be improved or tuned better. It would also be interesting to construct multivariate KDE's over the configuration space to estimate density better and make better estimates of uncertainty for unseen candidate configurations caused by sparsity in the distribution of observed configurations.


With regards to the curse of dimensionality in querying a kD-Tree, we suggested using Principle Component Analysis, but there are other options such as Locality-Sensitive Hashing to explore and experiment with.

\bibliographystyle{splncs04.bst}
\bibliography{paper.bib}
\renewcommand\thefigure{A.\arabic{figure}} 
\renewcommand\thetable{A.\arabic{figure}} 
\newpage
\section{Appendix}
\setcounter{figure}{0}

\begin{table}[]
\centering
	\begin{tabular}{lllrr}
		\hline
		\textbf{task} & \textbf{dataset name}            & \textbf{\# samples} & \textbf{\# features}\\ \hline
		3             & kr-vs-kp                         & 3196                & 37                   \\
		15            & breast-w                         & 699                 & 10                   \\
		29            & credit-approval                  & 690                 & 16                   \\
		31            & credit-g                         & 1000                & 21                   \\
		37            & diabetes                         & 768                 & 9                    \\
		43            & spambase                         & 4601                & 58                   \\
		49            & tic-tac-toe                      & 958                 & 10                   \\
		219           & electricity                      & 45312               & 9                    \\
		3021          & sick                             & 3772                & 30                   \\
		3902          & pc4                              & 1458                & 38                   \\
		3903          & pc3                              & 1563                & 38                   \\
		3904          & jm1                              & 10885               & 22                   \\
		3913          & kc2                              & 522                 & 22                   \\
		3917          & kc1                              & 2109                & 22                   \\
		3918          & pc1                              & 1109                & 22                   \\
		7592          & adult                            & 48842               & 15                   \\
		9946          & wdbc                             & 569                 & 31                   \\
		9952          & phoneme                          & 5404                & 6                    \\
		9957          & qsar-biodeg                      & 1055                & 42                   \\
		9971          & ilpd                             & 583                 & 11                   \\
		9978          & ozone-level-8hr                  & 2534                & 73                   \\
		10093         & banknote-auth...          & 1372                & 5                    \\
		10101         & blood-transfusion... & 784                 & 5                    \\
		14952         & PhishingWebsites                 & 11055               & 31                   \\
		14954         & cylinder-bands                   & 540                 & 40                   \\
		14965         & bank-marketing                   & 45211               & 17                   \\
		125920        & dresses-sales                    & 500                 & 13                   \\
		146818        & Australian                       & 690                 & 15                   \\
		146819        & climate-model... & 540                 & 21                   \\
		146820        & wilt                             & 4839                & 6                    \\
		167141        & churn                            & 5000                & 21                   \\\hline\\
	\end{tabular}
	\caption{Overview of tasks used for experiments with LightGBM configurations.}
	\label{tab:task-overview-lightgbm}
\end{table}

\begin{figure}
  \begin{subfigure}{\linewidth}
  \includegraphics[width=0.5\linewidth]{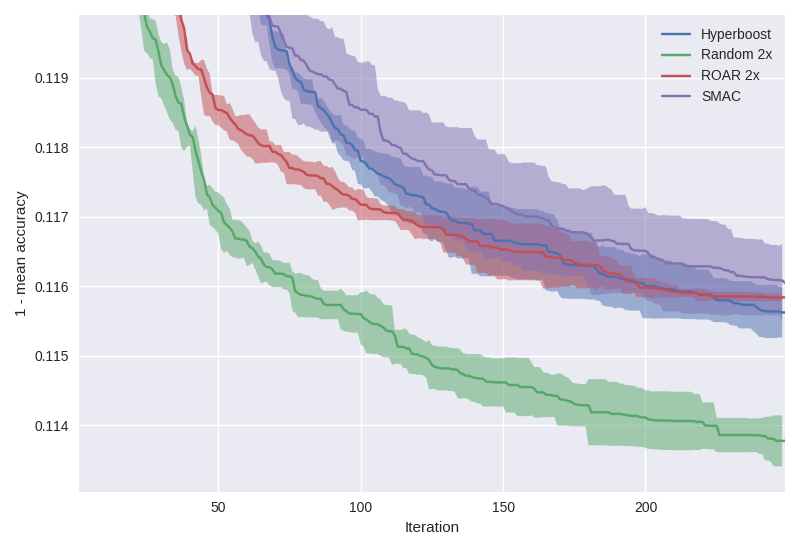}\hfill
  \includegraphics[width=0.5\linewidth]{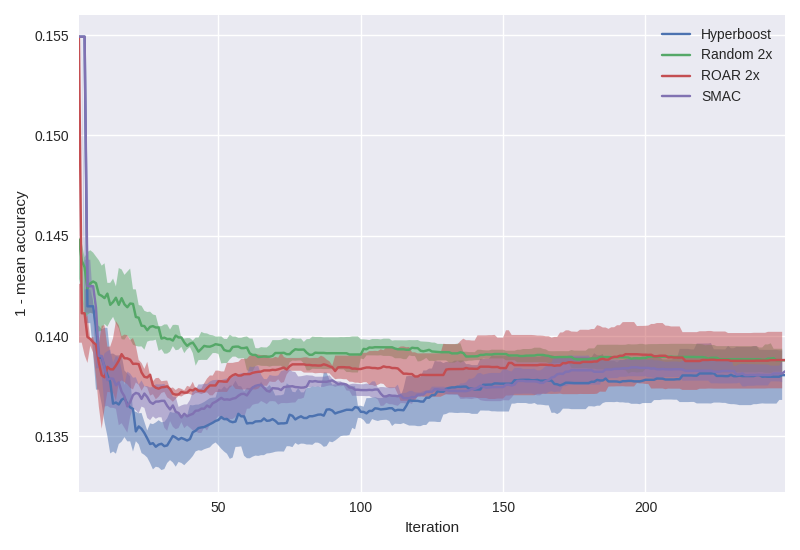}\hfill
  \caption{Mean loss on training set (left) and test set (right) for optimizing the Decision Tree target model}
  \label{fig:dt-mean}
  \end{subfigure}\par\medskip
  \begin{subfigure}{\linewidth}
  \includegraphics[width=0.5\linewidth]{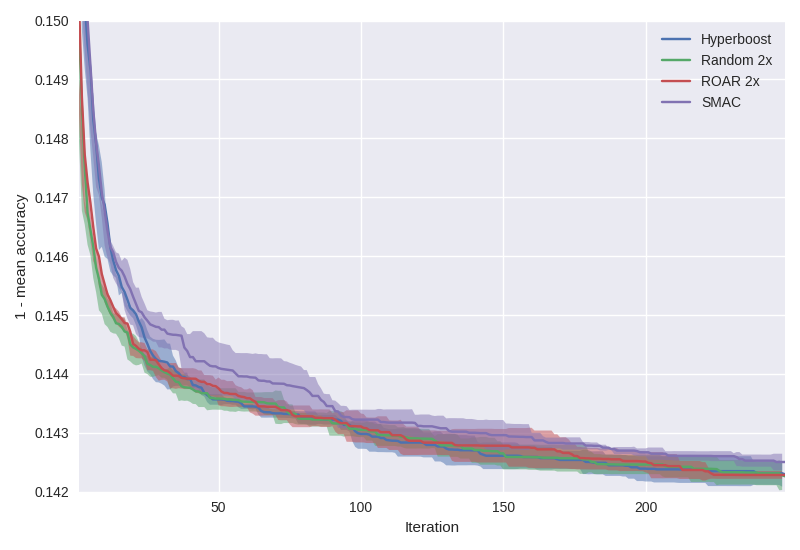}\hfill
  \includegraphics[width=0.5\linewidth]{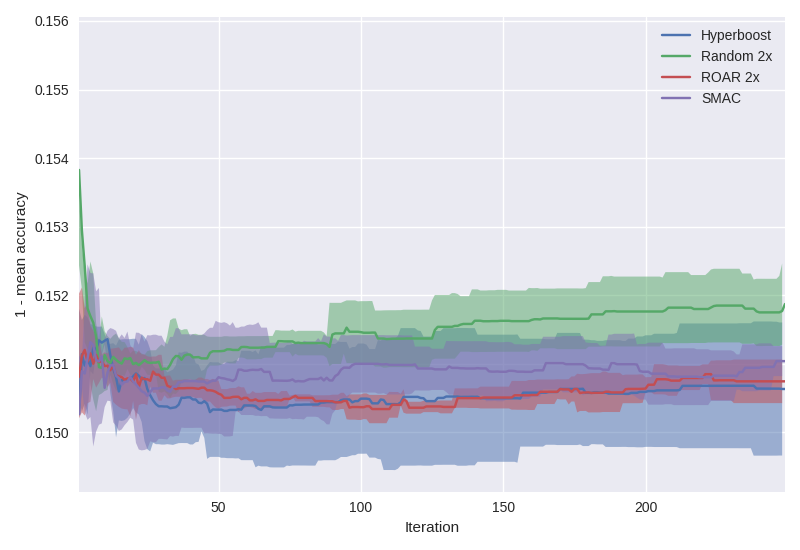}\hfill
  \caption{Mean loss on training set (left) and test set (right) for optimizing the SVM target model}
  \label{fig:svm-mean}
  \end{subfigure}\par\medskip
  \begin{subfigure}{\linewidth}
  \includegraphics[width=0.5\linewidth]{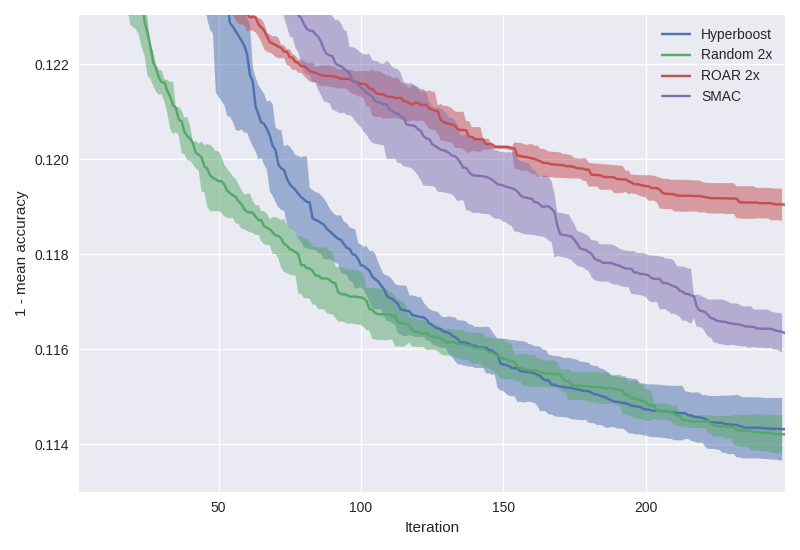}\hfill
  \includegraphics[width=0.5\linewidth]{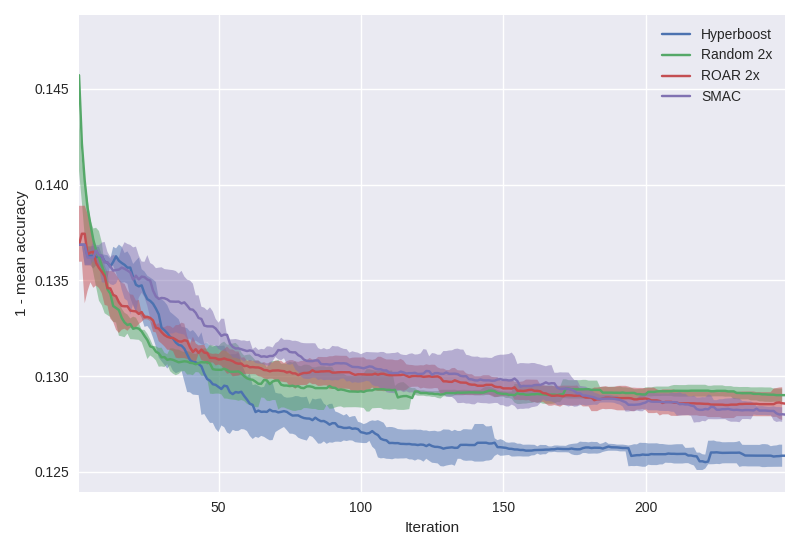}\hfill
  \caption{Mean loss on training set (left) and test set (right) for optimizing the Random Forest target model}
  \label{fig:rf-mean}
  \end{subfigure}
\end{figure}

\end{document}